\documentclass{article}

\usepackage{microtype}
\usepackage{graphicx}
\usepackage{subfigure}
\usepackage{booktabs} 
\usepackage{hyperref}

\usepackage[accepted]{icml2022}
\usepackage{amsmath}
\usepackage{amssymb}
\usepackage{mathtools}
\usepackage{amsthm}
\usepackage[capitalize,noabbrev]{cleveref}

\theoremstyle{plain}

\theoremstyle{definition}

\theoremstyle{remark}

\usepackage[textsize=tiny]{todonotes}

\icmltitlerunning{Professional Agents}

\begin{document}

\twocolumn[
\icmltitle{Professional Agents - Evolving Large Language Models into Autonomous Experts with Human-Level Competencies}

\begin{icmlauthorlist}
\icmlauthor{Zhixuan Chu}{ant}
\icmlauthor{Yan Wang}{ant}
\icmlauthor{Feng Zhu}{ant}
\icmlauthor{Lu Yu}{ant}
\icmlauthor{Longfei Li}{ant}
\icmlauthor{Jinjie Gu}{ant}
\end{icmlauthorlist}

\icmlaffiliation{ant}{Ant Group, Hangzhou, China}
\icmlcorrespondingauthor{}{{chuzhixuan.czx, luli.wy, zhufeng.zhu, bruceyu.yl, longyao.llf, jinjie.gujj}@antgroup.com}
\icmlkeywords{Machine Learning, ICML}

\vskip 0.3in
]

\printAffiliationsAndNotice{}

\begin{abstract}
\vspace{-1mm}
The advent of large language models (LLMs) such as ChatGPT, PaLM, and GPT-4 has catalyzed remarkable advances in natural language processing, demonstrating human-like language fluency and reasoning capacities. This position paper introduces the concept of Professional Agents (PAgents), an application framework harnessing LLM capabilities to create autonomous agents with controllable, specialized, interactive, and professional-level competencies. We posit that PAgents can reshape professional services through continuously developed expertise. Our proposed PAgents framework entails a tri-layered architecture for genesis, evolution, and synergy: a base tool layer, a middle agent layer, and a top synergy layer. This paper aims to spur discourse on promising real-world applications of LLMs. We argue the increasing sophistication and integration of PAgents could lead to AI systems exhibiting professional mastery over complex domains, serving critical needs, and potentially achieving artificial general intelligence.

\end{abstract}
\vspace{-5mm}
\section{Introduction}
\vspace{-2mm}
The emergence of large language models (LLMs) such as GPT-3 \cite{gpt-3-conf/nips/BrownMRSKDNSSAA20}, PaLM \cite{PaLM-journals/jmlr/ChowdheryNDBMRBCSGSSTMRBTSPRDHPBAI23}, and ChatGPT signifies a pivotal milestone in the advancement of artificial intelligence. These models leverage the sheer scale of data and computational power to display impressive language fluency, textual cohesion, and logical reasoning capabilities \cite{LLMRec-journals/corr/abs-2308-10837,LLMRG-journals/corr/abs-2308-10835,LLMHG-journals/corr/abs-2401-08217,jin2023time,xue2023prompt}. LLMs can craft high-quality long-form text, hold conversations, translate between languages, comprehend and respond to queries, summarize lengthy documents, and even generate code based on textual descriptions \cite{InstructGPT-conf/nips/Ouyang0JAWMZASR22,Wenxiang-conf/emnlp/Jiao0W0LW0T23,Hannah-conf/emnlp/KirkBVRH23}. The incremental improvement displayed with each successive LLM model release validates the potential as a viable path forward for artificial general intelligence (AGI). 

Essentially, they display signs of understanding natural language, mimicking capacities once considered exclusive to human intelligence. Therefore, some academics envision a future where they parallel or even surpass human aptitude across numerous cognitive tasks \cite{ERLHF-journals/corr/abs-2309-00754, RLAIF-journals/corr/abs-2309-00267}. Some theorists speculate that adequately advanced LLMs could autonomously persist self-improvement, acquiring greater-than-human reasoning and communication skills through an iterative process of self-updating their parameters by ingesting more information \cite{ReST-ReAct-journals/corr/abs-2312-10003, CIF-conf/acl/ToBGN23, RIF-journals/corr/abs-2305-14483, selfrew-journals/corr/abs-2401-10020}. It is worth exploring that LLMs can demonstrate abilities similar to or exceeding humans in different vertical areas \cite{MolXPT-conf/acl/LiuZXW0QZL23, SMILES-BERT-conf/bcb/WangGWSH19, MlPS-conf/iclr/WangZHYJGC23, DCLLM-journals/corr/abs-2310-17784}. Additionally, the forms in which LLMs will manifest such capabilities merit further investigation.

Agents, defined as artificial intelligence systems capable of autonomous planning and action to achieve specified goals, have long shown promise as a pathway to achieving artificial general intelligence \cite{agent-1-conf/nips/DevidzeRKS21,agent-3-conf/emnlp/0001YMWRCYR23,agent-2-conf/nips/YangLFS0Z20}. The capability of agents to self-direct their planning and learning processes based on high-level objectives is what may ultimately enable AGI \cite{Zhibin-journals/corr/abs-2305-11738, Zhenhailong-journals/corr/abs-2307-05300, Shaohui-journals/corr/abs-2309-01352}. This would allow an intelligent agent to adaptively acquire the vast array of skills and knowledge displayed by human experts in fields like science, engineering, medicine, commerce, and more \cite{Ying-journals/tcbb/ZhangGLYSY23,Erik-journals/corr/abs-2206-13517,Gengmo-conf/iclr/ZhouGDZXWZK23}. With enough learning over time and experience, agents could match or surpass the best human specialists. The advent of agents with human and super-human level AGI promises to transform professional services in revolutionary ways. Intelligent agents could take on many of the analytical, creative, and decision-making duties currently performed by people \cite{Siqi-conf/emnlp/Ouyang023,Siqiao-journals/corr/abs-2308-05361,Liang-journals/corr/abs-2310-02071, LLMPA-journals/corr/abs-2312-06677}. This may greatly expand the capacities of companies and industries, allowing them to operate with unprecedented speed, quality, and scope of services. Agents with AGI therefore have immense potential economic and societal impact. 

In conclusion, we put forward Professional Agents (PAgents) as a promising pathway for achieving artificial general intelligence in the real world, which autonomously constructs and evolves agents to acquire and demonstrate professional-level competencies. It has a sophisticated tri-layered framework including the base tool layer, middle agent layer, and top synergy layer. PAgents are initialized according to the specified ``gene'', such as professional role, key tasks, and responsibilities, relevant datasets and corpora, as well as specifics of the knowledge domain. Then, they can constantly evolve by self-evolution, coevolution, human feedback evolution, gene refinement, and guidance from superstratum PAgents. Finally, groups of PAgents can synergistically collaborate by forming multi-agent systems that leverage individual strengths, account for conflicts, and work interdependently towards common goals in shared environments. Overall, the development of PAgents promises to push LLMs to new frontiers in replicating multifaceted human intelligence. 

\vspace{-2mm}
\section{Agent Examples} 
\vspace{-2mm}
This section will delve into a selection of existing agents, highlighting how each exemplifies core capabilities that contribute to the broader vision of professional agents.

\textbf{AutoGen} \cite{wu2023autogen} is a generalized multi-agent conversation framework. Crucially, all agents are made conversable - they can receive, react to, and respond to messages. When configured properly, agents can have multi-turn conversations with minimal human input, enabling both automation and human agency. AutoGen can use a ``conversation programming'' paradigm to unify complex LLM workflows as agent interactions.

\textbf{AppAgent} \cite{yang2023appagent} is a framework designed to operate smartphone apps like human users. It learns by first exploring apps on its own, interacting through predefined actions, and learning from the outcomes. These interactions are documented, helping AppAgent navigate the apps. This autonomous learning process can be sped up by observing some human demonstrations. After this exploration, AppAgent can consult the documentation to operate apps based on their current state, without needing extensive app-specific training data or adapting language model parameters.

\textbf{MetaGPT} \cite{hong2023metagpt} is a meta-programming framework for multi-agent collaboration. A key innovation of MetaGPT is integrating human-like workflows throughout its design, which significantly enhances robustness and reduces unproductive agent interactions. It encodes Standardized Operating Procedures into prompt sequences for more streamlined collaboration between agents with human-like domain expertise. This allows agents to verify intermediate results and reduce errors.

\textbf{ModelScope-Agent} \cite{li2023modelscope} is designed for easy integration with model APIs and common APIs, offering a streamlined approach for leveraging AI. It offers a default tool library that accommodates a wide range of AI model APIs in domains such as natural language processing, computer vision, audio, and multi-modal fields, alongside extensive common APIs like search engines. Users can also add their custom API plugins and utilize an automatic retrieval system from the extensive tool library. 

\textbf{AutoAgents} \cite{chen2023autoagents} is designed to create and manage a team of specialized AI agents to execute diverse tasks effectively. The framework emphasizes continuous self-improvement of individual agents as well as joint enhancement to foster skill development and facilitate knowledge transfer within the team. It also introduces an Action Observer, an agent responsible for facilitating the effective distribution of tasks, information exchange, decision-making, and adaptation to dynamic environments.

\textbf{OpenAgents} \cite{xie2023openagents} is an open-source platform designed to bridge the gap between language agents and everyday users. It comprises three specialized agents: a Data Agent for Python and SQL data analysis, a plugin agent hosting over 200 API tools, and a Web Agent designed for autonomous web surfing. The platform caters to both general users and developers, furnishing a user-friendly web interface that allows non-experts to leverage agent functionalities without programming skills. 

\textbf{ChatDev} \cite{qian2023communicative} is a virtual assistant modeled after the traditional waterfall development method. It orchestrates the workflow through four distinct phases: design, coding, testing, and documentation. These phases are managed by ``software agents'', simulating roles such as programmers and testers, who collaborate through dialogue. 

\textbf{AGENTVERSE} \cite{chen2023agentverse} is a general multi-agent framework to simulate human group problem-solving and dynamically adjust group composition in response to ongoing progress. It divides the problem-solving process into four key stages: expert recruitment, collaborative decision-making, action execution, and evaluation. This allows for iterative refinement based on feedback. 

While these examples showcase preliminary implementations of professional agents, they demonstrate the types of capabilities and basic architecture. The next section will summarize key features and gaps seen across these examples to construct a comprehensive framework for developing full-fledged professional agents.

\begin{figure*}[t]
    \centering
    \includegraphics[width=1.82\columnwidth]{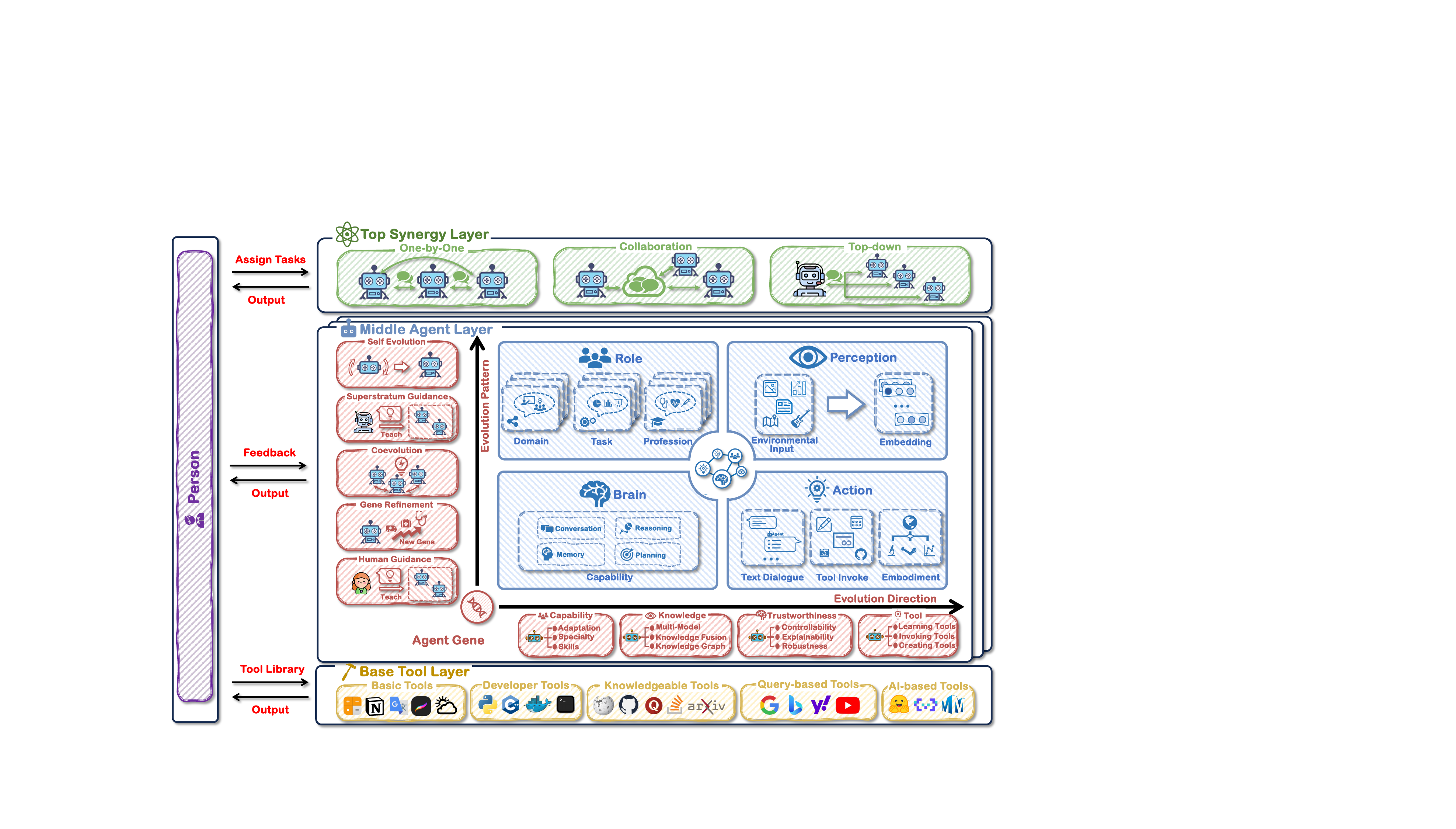}
    \vspace{-3mm}
    \caption{The standard PAgent has a general framework, i.e., a sophisticated tri-layered framework. Specifically, individual PAgent features a modular architecture consisting of role, perception, brain, and action components.}
    \vspace{-5mm}
    \label{fig:framework}
\end{figure*}

\vspace{-2mm}
\section{Framework of Professional Agents}
\vspace{-2mm}
The professional agent stands as a pioneering innovation in the realm of artificial intelligence, purpose-built to emulate the remarkable abilities of human professionals in continuous learning and adaptability. As shown in Figure \ref{fig:framework}, the standard PAgent has a general framework, i.e., \textbf{\textit{a sophisticated tri-layered framework}}:
\vspace{-3mm}
\begin{itemize}
\setlength\itemsep{-0.2em}
\item  \textit{Base Tool Layer}: This foundational layer provides a wide range of state-of-the-art technologies, spanning basic utilities, developer resources, knowledge repositories, search capabilities, and AI systems. This diverse set of tools lays the groundwork for the infrastructure, facilitating the emergence and continuous improvement of skilled agents.

\item  \textit{Middle Agent Layer}: There is a diverse collection of independent, controllable, interactive, specialized professional agents. Each PAgent embodies a particular professional role and the key competencies needed to fulfill that role effectively. 

\item  \textit{Top Synergy Layer}: This is a harmonious fusion of individual professional agents into an integrated network. In concert, these agents pool their expertise to address multifaceted tasks, emulating the cooperation inherent in human professional groups. They dynamically allocate tasks, adjusting fluidly to the varying complexity and contextual demands that arise.
\end{itemize}

\vspace{-2mm}
\section{Components of PAgents}\label{sec.pagent}
\vspace{-2mm}
Professional agents are artificial intelligence systems designed to acquire specialized expertise and demonstrate professional-level competencies like human professionals. The representative professional agents have four modules, i.e., \textbf{\textit{Role}}, \textbf{\textit{Perception}}, \textbf{\textit{Brain}}, and \textbf{\textit{Action}}. 

\vspace{-2mm}
\subsection{Role}
\vspace{-2mm}
The role module is a key component of the PAgent that defines the professional identity and capabilities that the agent will develop. The role module specifies metadata such as the profession, specialty, skills, credentials, and experiences of the agent. This foundational information guides the development and bounded growth of the perception, brain, and action modules. The role module connects to the \textit{perception} module by indicating the forms of sensory inputs that are relevant to the profession. To guide the \textit{brain} module, the role module translates professional competencies into concrete learning goals and curricula. A software engineer role module may decompose broad engineering capabilities into discrete modules focused on programming languages, debugging skills, framework knowledge, communication protocols, design methodologies, and more. The brain module leverages these curricular signals to focus its acquisition and reinforcement of the prescribed knowledge and skills. For the \textit{action} module, the role module characterizes appropriate behavioral patterns. By internalizing role directives, the action module gains purposeful direction to expand its tools to fulfill professional demands. The role module connects the PAgent architecture into an integrated system for actualizing specialized expertise. 

\vspace{-2mm}
\subsection{Perception} 
\vspace{-2mm}
The perception module is responsible for ingesting and encoding multimodal professional data that the PAgent receives as inputs. This module allows the PAgent to perceive complex real-world professional scenarios across diverse domains through multiple data formats including text, images, audio, and video. Specifically, the perception module consists of advanced encoders tailored to each modality: Text Encoder, Image Encoder, Audio Encoder, Video Encoder, and so on. Through these tailored encoders, the perception module can interpret complex real-world professional data into structured representations that the brain module can readily comprehend and process to make intelligent decisions. In the following section, we will introduce how to enable our PAgent to acquire multimodal perception capabilities for textual input, visual input, and auditory input.

\subsubsection{Textual Input}

Similar to humans, an agent powered by a LLM primarily relies on text to transmit data, information, and knowledge. The PAgent, built on LLMs, fundamentally communicates with humans through text input and output. However, understanding the implicit meanings in certain texts remains a significant challenge for LLM-based agents, as these subtleties can be difficult to interpret. Some existing works \cite{christiano2017deep,basu2018learning,sumers2021learning,lin2022inferring} have attempted to address this issue to learn the implied meanings from textual feedback. The user's preferences can be inferred to enable the agent to deliver more tailored and precise responses.

\subsubsection{Visual Input}

As a language, LLMs have shown their outstanding capability in understanding languages \cite{achiam2023gpt} and conversations \cite{lu2020improving}. However, some of the existing LLM-based agents still lack a transformation ability from images to textual content. Images from the agent's surroundings always contain a wealth of information about objects, spatial relationships, layouts, and so on. Therefore, integrating the image information into our PAgent is the first step to achieving multimodal perception. The existing works in this research direction can be generally classified into three categories: (1) image captioning, (2) image representation, and (3) video understanding.

\textit{Image Captioning.} This is a straightforward way to understand image inputs, which is applied to generate the corresponding text descriptions for the images \cite{vaswani2017attention,chen2021visualgpt,touvron2021training}. One of the most significant advantages of this approach is that captions can be easily linked with structured instructions and fed into the agent. However, it inevitably loses a lot of potential information during the conversion process \cite{driess2023palm}. Also, based on this approach, the agent may introduce some biases. \textit{Image Representation.} Transformers have been applied in the field of computer vision to represent images. For example, ViT/VQVAE \cite{van2017neural,dosovitskiy2020image,mehta2021mobilevit} are proposed to encode visual information by using transformers. Also, some approaches with pure MLP structures such as Mlp-mixer \cite{tolstikhin2021mlp} achieve the excellent performance of image representation. Some pre-trained visual encoders can enhance the agent's visual perception \cite{xi2023rise}. \textit{Video Understanding.} Compared with images, videos add an additional temporal dimension. Therefore, it is crucial for the agent to recognize the relationships among a series of continuous image frames. Some existing works such as Flamingo \cite{alayrac2022flamingo} tend to ensure a temporal order when understanding videos. However, these approaches usually use a masking mechanism and thus they may restrict the agent's view and lose some potential image information.

\subsubsection{Auditory Input}
Apart from textual and visual information, auditory information is a crucial component of world information as well. The existing agents tend to use well-established models \cite{huang2023audiogpt,ren2019fastspeech,ye2022syntaspeech} as tools to receive auditory information. For example, AudioGPT \cite{huang2023audiogpt} leverages the strengths of advanced models including FastSpeech \cite{ren2019fastspeech}, GenerSpeech \cite{ye2022syntaspeech}, Whisper \cite{ye2022syntaspeech}, and various others \cite{kim2021conditional,wang2023tf}, which have demonstrated remarkable success in tasks like Text-to-Speech conversion, Style Transfer, and Speech Recognition.

\vspace{-2mm}
\subsection{Brain}
\vspace{-2mm}
Much like humans, the brain of our PAgent serves as a central nucleus driven by an LLM. The brain module enables PAgents to demonstrate sophisticated cognitive abilities critical for professional-grade performance. Built upon LLMs, the brain module encompasses four key capabilities: \textit{\textbf{Conversation}}, \textit{\textbf{Memory}}, \textit{\textbf{Planning}}, and \textit{\textbf{Reasoning}}.

\subsubsection{Conversation}
With the help of powerful LLMs \cite{workshop2022bloom,touvron2023llama}, the PAgent can easily interact with other agents, i.e., communication among multiple agents. Also, the LLMs allow users to understand the response and reaction of the PAgent \cite{serban2016generative,adiwardana2020towards}. The LLMs can provide multi-turn interactive conversation \cite{chung2022scaling,touvron2023llama}, high-quality natural language generation \cite{bang2023multitask}, and implication understanding \cite{shapira2023clever} capabilities.
\subsubsection{Memory}
The memory mechanism can store the knowledge learned from the real world and its historical behaviors, e.g., observations, thoughts, and actions. Similar to the processes of human strategy formulation and decision-making, the PAgent also uses its memory mechanism to handle a sequence of elaborate tasks. The memory mechanism includes the following three parts, i.e., structures, formats, and operations.
\vspace{-5mm}
\begin{itemize}
\setlength\itemsep{-0.2em}
    \item \textit{Memory Structures}: LLM-based autonomous agents are designed with memory structures inspired by human memory processes, as studied in cognitive science \cite{nuxoll2007extending}. These structures \cite{rana2023sayplan,madaan2022memory,modarressi2023ret,schuurmans2023memory,zhong2023memorybank,zhu2023ghost,shinn2023reflexion} mimic the progression from sensory memory to short-term memory, and finally to long-term memory. Short-term memory is reflected in the transformer architecture's context window that temporarily holds input information. Long-term memory parallels an external storage from which agents can quickly fetch and use information. The text goes on to present two memory structures that are modeled after these short- and long-term human memories.
    \item \textit{Memory Formats}: The memory formats for Agents can be generally classified into the four main categories, i.e., natural languages, embeddings, databases, and structured lists \cite{wang2023survey}. The format of natural languages tends to describe the world knowledge and the agent's behaviors in raw natural language, which can be easily understood by LLMs \cite{shinn2023reflexion}. The format of embeddings tends to store embedding vectors, which can enhance memory retrieval and reading capabilities \cite{zhong2023memorybank}. The memory stored in databases allows the agent to manipulate memories \cite{hu2023chatdb}.
    As for the final format, i.e., structured lists, the memory can be conveyed in an efficient and concise manner, e.g., JSON \cite{zhu2023ghost,modarressi2023ret}. 
    \item \textit{Memory Operations}: There are three crucial memory operations for the PAgent, i.e., reading, writing, and reflection. Memory reading is the ability to extract meaningful information from memory to assist the agent's actions. For example, learning from the previous actions to achieve similar goals \cite{zhu2023ghost}. Memory writing is used to store information from the surroundings or the agents in memory, which can handle the problems of memory duplication \cite{schulman2017proximal} and memory overflow \cite{modarressi2023ret}.
\end{itemize}
\subsubsection{Planning and Reasoning}
Humans tend to apply a kind of heuristic thinking to deconstruct a complex task into a sequence of subtasks and then complete the subtasks one by one. Therefore, the planning and reasoning component of the PAgent's brain is designed to perform as humans when faced with an elaborate task. The mainstreaming planning or reasoning strategies of the agents can be generally classified into two groups: (1) planning without feedback and (2) planning with feedback.
\vspace{-3mm}
\begin{itemize}
\setlength\itemsep{-0.2em}
    \item \textit{Planning without Feedback}: In this mode, the PAgent receives no feedback from users or its historical trials. This means that the historical actions of PAgent cannot influence its future planning. Many existing works are focussing on this mode. Here, we present three representative strategies, i.e., single-path reasoning, multi-path reasoning, and external planning. Single-path reasoning, e.g., Chain of Thought (COT) \cite{wei2022chain}, Zero-shot-COT \cite{kojima2205large}, ReWOO \cite{xu2023rewoo}, tend to decompose the final task into several intermediate steps. Multi-path reasoning tends to decompose a task into a tree-like structure, e.g., Self-consistent CoT (CoT-SC) \cite{wang2022self}, Tree of Thoughts (ToT) \cite{yao2023tree}, Graph of Thoughts (GoT) \cite{besta2023graph}, and Algorithm of Thoughts (AoT) \cite{sel2023algorithm}. External planning tends to leverage external tools to employ efficient search algorithms to identify correct plans, e.g., LLM+P \cite{liu2023llm+}, LLM-DP \cite{dagan2023dynamic}, and CO-LLM \cite{zhang2023building}.
    \item \textit{Planning with Feedback}: In this mode, the PAgent can make long-horizon planning to solve complex tasks by considering users' or its feedback. According to the sources of feedback, the existing approaches can be classified into three categories, i.e., environmental feedback, human feedback, and model feedback. Environmental feedback is obtained from the real world or virtual environment, e.g., thought-act-observation triplets (ReAct \cite{yao2022react}), execution progress\&error\&verification (Voyager \cite{wang2023voyager}), scene graph (SayPlan \cite{rana2023sayplan}), and planning feedback (LLM-Planner \cite{song2023llm}). Human feedback can help the PAgent align with human values and preferences, e.g., Inner Monologue \cite{huang2022inner}. Model feedback is obtained from the agents themselves, which can provide a kind of self-reflection mechanism, e.g., Self-refine \cite{madaan2023self}, Self-Check \cite{miao2023selfcheck}, and Reflexion \cite{shinn2023reflexion}. 
\end{itemize}

\vspace{-2mm}
\subsection{Action}
\vspace{-1mm}
The action component of the PAgent is designed to translate decisions into specific outcomes. The action module enables PAgents to demonstrate sophisticated professional capabilities by processing inputs received from the perception and brain modules. In this section, we introduce the action component from the following three perspectives: (1) action goal, (2) candidate actions, and (3) action production.
\vspace{-3mm}
\begin{itemize}
\setlength\itemsep{-0.2em}
    \item \textit{Action Goal}: Action goal is the objective of acting and it can be classified into different groups, such as task completion, communication keeping, and environment exploration.
    \item \textit{Candidate Actions}: These actions are collected into an action space and can be performed by the PAgent. The candidate actions can be generally classified into two groups, i.e., external tools (such as APIs \cite{chen2021visualgpt,schick2023toolformer,li2023api,qin2023toolllm}, databases \cite{hu2023chatdb}, and external models \cite{zhong2023memorybank}) and internal knowledge \cite{wang2023voyager,fischer2023reflective}.
    \item \textit{Action Production}: Unlike conventional LLMs, the PAgent can take actions with different strategies and sources, which means that action production should recollect memory \cite{park2023generative} and follow plans \cite{wang2023describe}.
\end{itemize}

\vspace{-4mm}
\section{Architecting Professional Agents}
\vspace{-1mm}
\subsection{Genesis}
\vspace{-1mm}
A key aspect of PAgents is the notion of a ``gene'' which encapsulates the core functionality, competencies, and knowledge required for the agent to operate effectively within its target domain. The gene constitutes descriptors related to the agent's professional role, key tasks and responsibilities, relevant datasets and corpora, as well as specifics of the knowledge domain. Based on this gene, a PAgent can essentially build itself by initializing its modular components to align with the specifications set forth. Below is a high-level Standard Operating Procedure (SOP) or pipeline for constructing and evolving a PAgent:
\vspace{-3mm}
\begin{itemize}

\setlength\itemsep{-0.2em}

\item  \textit{Define the Professional Role}: Identify the professional domain and the specific role the PAgent will fulfill. Determine the necessary skills, credentials, experiences, and professional competencies required for the role. Create a role module that encapsulates this information, using metadata, natural language descriptions, or executable logic to outline the agent's professional identity.

\item \textit{Develop the Perception Module}: Identify the types of sensory inputs relevant to the professional role, such as text, images, audio, or video. Implement advanced encoders tailored to each modality to interpret complex real-world professional data. Ensure the perception module can convert the diverse sensory data into structured representations for further processing.

\item \textit{Build the Brain Module}: Utilize large language models to facilitate natural language processing, knowledge acquisition, memory, reasoning, and planning. Define the learning goals and curricula based on the competencies outlined in the role module. Implement mechanisms for continuous learning to keep the PAgent up-to-date with the latest industry knowledge and practices.

\item \textit{Construct the Action Module}: Define the behavioral patterns required for the professional role, including tool utilization and potential physical embodiment. Integrate consultation and decision-making abilities to allow the PAgent to provide expert advice and perform tasks. Ensure the action module can execute sophisticated tasks, receive feedback, and adapt its strategies for improved performance.

\item \textit{Integration and Testing}: Integrate the Perception, Brain, and Action modules into a cohesive system. Test the PAgent in controlled environments, simulating real-world professional scenarios to evaluate its performance. Use feedback from testing to refine the agent's capabilities and address any deficiencies.

\item \textit{Deployment}: Deploy the PAgent in a real-world professional setting, initially under supervision to monitor its performance and impact. Establish protocols for human-agent interaction, ensuring that the PAgent can work collaboratively with human professionals.

\item \textit{Monitoring and Evolution}: Continuously monitor the PAgent's performance using key performance indicators relevant to the professional domain. Collect data on the PAgent's decisions, actions, and learning progress. Update the role module to reflect changes in the professional landscape or to expand the PAgent's capabilities.

\item \textit{Continuous Learning and Adaptation}: Implement feedback loops that allow the PAgent to learn from its successes and failures. Periodically review the latest research and technological advancements to update the PAgent's modules. Conduct ongoing training and re-training to ensure the PAgent maintains a high level of expertise and adapts to new information or procedures.

\item \textit{Ethical Considerations and Compliance}: Ensure that the PAgent adheres to ethical guidelines and professional standards specific to the domain it operates in. Integrate checks and balances to monitor for biases, errors, and compliance with legal and regulatory requirements.

\item \textit{Feedback and Iteration}: Gather feedback from end-users and stakeholders to gauge the efficacy and acceptability of the PAgent. Iterate on the design and functionalities of the PAgent based on user feedback, performance metrics, and evolving professional needs.
    
\end{itemize}
\vspace{-3mm}
This SOP is general and should be customized for each specific implementation of a PAgent. The complexity and requirements of the process will vary depending on the professional domain and the intended scope of the PAgent's role within that domain.

\vspace{-2mm}
\subsection{Evolution}
\vspace{-2mm}
Realizing an autonomous replication and adaptation agent requires a sophisticated and dynamic AI architecture that can evolve, much like a human professional advancing through different stages of their career. This entails evolution by five key patterns:
\vspace{-3mm}
\begin{itemize}
\setlength\itemsep{-0.2em}
\item \textit{Self-evolution:} The self-evolution capacity of PAgents relies on memory and learning mechanisms to accumulate knowledge and skills from experiences. As they take on more projects, the PAgents assimilate new information, tools, and best practices into their structure, constantly enhancing their competencies. Agents can refine their skills using three primary methods. Firstly, they employ memory components to record and recall data from past exchanges and guidance for future actions, especially recalling prior successes when facing similar objectives, as described by \cite{Wang2023ASO}. Secondly, agents evolve autonomously, going beyond memory to actively modify their aims and tactics, train based on interactions, and adapt to dynamic settings. This evolution is illustrated by \cite{Nascimento2023SelfAdaptiveLL} with a self-management protocol for agent adaptability and by \cite{Zhang2023ProAgentBP} with ProAgent, which adjusts strategies in response to team communications. \cite{Wang2023AdaptingLA} discusses a communication approach, utilizing agent dialogues as training data to enhance agents' learning beyond in-context learning or supervised fine-tuning. Lastly, dynamic generation allows the creation of new agents as needs arise during operations, providing tailored solutions to immediate challenges \cite{Chen2023AutoAgentsAF}.

\item \textit{Coevolution:} PAgents can mutually evolve through collaboration with peer agents. As diverse PAgents work together on complex tasks, they can learn from each other's specialized expertise areas, exchange feedback on effectiveness, and integrate complementary skills. This coevolution process iterates, allowing the PAgents to mature rapidly.

\item \textit{Human-in-the-loop Guidance:} Human experts further guide the development trajectory of PAgents by providing feedback, corrections, and new knowledge. With human-in-the-loop supervision, the agents steer their growth towards configurations that demonstrate higher quality outcomes and judgment expected by professionals. The integration of human feedback is crucial to the PAgent system's iterative refinement, fostering a learning loop that benefits from human expertise and ethical standards. The agents' decisions and actions are evaluated by human professionals, ensuring alignment with human values and professional norms.

\item \textit{Gene Refinement:}
The PAgent creators also facilitate systematic evolution by revisiting, removing, and adding new elements to the PAgents' fundamental genes. Adjusting the building blocks of their roles, responsibilities, tools and underlying knowledge domains gives rise to new and improved generations of PAgents. For example, task expansion needs agents to absorb or invent new capabilities, thus stretching to meet the broadening horizons of their roles. Domain transfer requires agents to adapt and thrive across various professional environments.

\item \textit{Guidance from Superstratum PAgents.}
Finally, the network effects produced by interactions between PAgents, especially with superstratum PAgents - more senior, vastly experienced agents - multiply their learning pace. The influx of macro-level guidance, project delegation, and exposure to multifaceted problem-solving shapes well-rounded, leadership-grade PAgents.

\end{itemize}

In addition to these five patterns, there are four key directions for the evolution of capabilities. For the role module, as PAgents take on more diverse professional tasks across different domains, the role definitions must evolve to encompass a wider range of capabilities. More complex, multi-faceted roles will require integrating distinct competencies and skills tailored to niche specializations. Dynamic role updating is necessary for flexibly taking on new challenges. For the perception module, advanced neural perception systems will empower PAgents to interpret inputs spanning text, imagery, audio, video, and more. Tailored modality encoders, developed using cutting-edge machine learning techniques, will enable nuanced encoding of intricate real-world data. For the brain module, incorporating mechanisms for causality, explainability, controllability, safety, accuracy, robustness, and ethical alignment will be critical for establishing trustworthiness \cite{chu2023task,guo2023fair,li2023machine,zhu2023trustworthy,chu2021graph}. External auditing of reasoning, allowing human override of decisions, sensitivity analysis, adversarial testing, and alignment techniques can help assure reliable and ethical intelligent behavior. For the action module, just as technological innovations have historically expanded human potential, the PAgent system maintains its professional agents at the cutting edge by integrating the latest AI methodologies and tools. This paves the way for more sophisticated forms of analysis, inference, and strategic planning.

\vspace{-2mm}
\subsection{Multi-agent Synergy}
\vspace{-2mm}
The development of a single agent usually sticks to its profile and excels at specialized skills \cite{Sumers2023CognitiveAF}. Such design gives them the professional ability to react to their surroundings according to particular internal settings. In contrast, an LLM-based Multi-Agent System (LLM-MAS) \cite{dorri2018multi} consists of multiple agents that can collaborate to accomplish a common or conflicting goal in the sharing environment. In comparison to conventional single-agent systems, multi-agent systems have several merits. Primarily, they excel in scalability, enabling them to tackle intricate tasks more effectively. Secondly, their inherent resilience allows them to withstand the malfunctioning or failure of individual components without collapsing. Lastly, by leveraging the distinct strengths and differences among agents, these systems can optimize problem-solving processes and enhance efficiency.

The above benefits make us believe that connecting the single PAgent to establish a Multi-PAgent System (MPAS) can extend the scope of capabilities for users to accomplish more complicated problems. Despite the advantages, we should pay attention to the challenges coexisted with the opportunities. First and foremost, the increasing number of agents raises great challenges in controlling them to communicate with each other to complete the given goal. In addition, the tasks allocated to the agent might have conflicts with each other, which needs careful design on how to coordinate them. Finally, agents may be required to adapt to learn from the dynamic environments. If we can appropriately handle the challenges, multiple agents offer a powerful approach to solving complex tasks.

Here, we clarify how communication is organized to coordinate different agents \cite{taicheng2024masSurvey}. Analogous with the other multi-agent systems, the essence of MPAS lies in fostering effective communication among agents. We explore this aspect through three lenses: (1) \emph{Communication Approaches}, which delves into the diverse manners and techniques employed by agents to communicate with one another; (2) \emph{Communication Network Organization}, focusing on how the communication network within a multi-agent system is structured and designed for smooth information flow; (3) \emph{Agent Message Exchange}, referring to the information being shared between agents during their interactions, which forms a cornerstone of their collaborative efforts.

\begin{itemize}
\setlength\itemsep{-0.2em}
\item \textit{Communication Approaches:} Presently, the predominant frameworks for interaction within LLM-MA systems are categorized as \emph{Cooperative}, \emph{Debate}, and \emph{Competitive} models. In the Cooperative model, agents collaborate and aim for a common purpose or targets, sharing information to improve the overall outcome. When agents partake in discussions involving critical exchanges, the Debate model comes into play, where they advocate for and defend their individual perspectives or proposals, while also evaluating and challenging those of their peers. This approach is particularly effective for forging agreement or enhancing a solution. In the competitive model, agents pursue their objectives, which may not align and could potentially clash with the aspirations of other participants.

\item \textit{Communication Network Organization:} Apart from the communication approaches, the way to organize the agents directly has an impact on communication efficiency and performance. Generally speaking, the communication architecture mainly includes a hierarchical layer, ego-network with a central master node, decentralized network, and shared message pool. In a hierarchical layered system, agents are organized by tiers, each playing specific roles and interacting mostly within their level or with those immediately above or below. The Dynamic LLM-Agent Network (DyLAN), as presented by \cite{Liu2023DynamicLN}, arranges agents across multiple layers in a directed feed-forward architecture. This enables dynamic exchanges with capabilities such as on-the-fly agent choice and an early termination feature to improve collaborative effectiveness. Meanwhile, decentralized communication is characterized by a direct, peer-to-peer dialogue among agents. In contrast, centralized ego-network communication revolves around one or several pivotal agents orchestrating the network's interactions, in which subsidiary agents are linked through a core hub. To elevate communication efficacy, MetaGPT \cite{Hong2023MetaGPTMP} introduces a shared message pool. In this system, a universal pool of messages is maintained for agents to post updates and subscribe to messages, through which the communication process flows back and forth.

\item \textit{Agent Message Exchange:} In the arena of LLM-MA systems, textual data is commonly the vehicle for conveying messages. This textual data is diverse and tailored to suit the needs of the specific use case. For instance, within the software creation sphere, agents often exchange insights related to blocks of code. Conversely, during game simulations such as those based on Werewolf, dialogue between agents may revolve around their deductive reasoning, doubts they harbor, or planned methodologies. Expanding upon this, it's conceivable that in more nuanced applications, agents might also share semantic cues or emotional intent through text, which could be critical in contexts like customer service or interactive storytelling, where understanding sentiment and context is paramount.

\end{itemize}

\section{Discussion}

This position paper introduces the concept of Professional Agents (PAgents), an application framework that harnesses the capabilities of large language models to create autonomous agents with specialized expertise. The genesis, evolution, and multi-agent synergy aspects offer a pipeline for constructing PAgents that can replicate and exceed human-level expertise through continuous learning. However, there remain significant research challenges to fully realizing professional mastery in PAgents across complex vocations. Key limitations include sample inefficiency in learning, brittleness when operating outside training distributions, insufficient reasoning abilities for multifaceted tasks, expressing complex motor skills for physical embodiment, and difficulty ensuring human-aligned behavior as agent capabilities grow. Mitigating these issues to create increasingly capable, safe, and robust PAgents will necessitate innovations in prompt engineering, simulator design, memory architectures, model training techniques, explainability methods, and policy learning algorithms.

\newpage
\bibliography{pagent}
\bibliographystyle{icml2024}

\end{document}